\documentclass[10pt,twocolumn,letterpaper]{article}

\usepackage[pagenumbers]{cvpr} 

\definecolor{cvprblue}{rgb}{0.21,0.49,0.74}
\usepackage[pagebackref,breaklinks,colorlinks,allcolors=cvprblue]{hyperref}
\usepackage{multirow}
\usepackage{colortbl}
\definecolor{oursblue}{HTML}{E3EEF9}
\usepackage{pifont}

\newcommand{\cnum}[1]{%
  \raisebox{-0.5pt}{\ding{\the\numexpr181+#1\relax}}%
}

\def\confName{CVPR}
\def\confYear{2027}

\newcommand{\sigt}{\sigma_\theta}
\newcommand{\ext}{E}

\title{CAT-Free: Calibration-, Annotation-, and Target-Training-Free Multi-View BEV Pedestrian Detection and Tracking}

\author{
Taigo Sakai$^{1}$\quad Hiroki Kouno$^{2}$\\
Naoki Kato$^{2}$\quad Kazuhiro Hotta$^{1}$\\
$^{1}$Meijo University 
    \qquad $^{2}$Chubu Electric Power Co., Inc.\\
{\tt\small 263441505@ccmailg.meijo-u.ac.jp, Kouno.Hiroki@chuden.co.jp,}\qquad\\
{\tt\small  Katou.Naoki7@chuden.co.jp, kazuhotta@meijo-u.ac.jp}
}

\begin{document}
\twocolumn[{%
\renewcommand\twocolumn[1][]{##1}%
\maketitle
\begin{center}
  \includegraphics[width=0.85\textwidth]{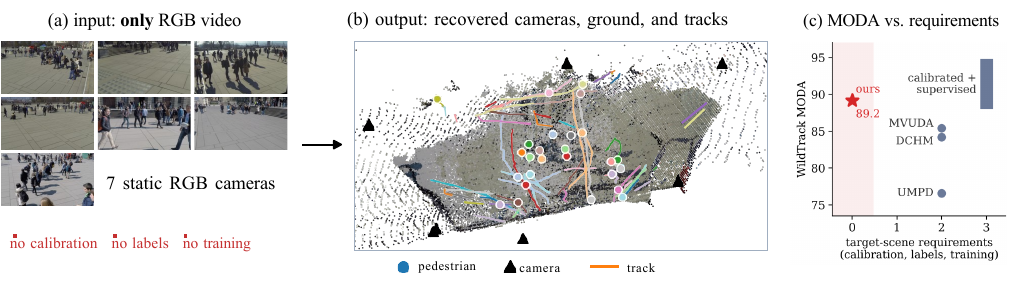}
  \captionof{figure}{Overview of CAT-Free on WildTrack.
  (a) The only scene-specific input is synchronized RGB video from 7 static cameras. Known camera parameters, position annotations from the target scene, or target-scene training are {NOT} used.
  (b) Cameras, ground, and all pedestrian trajectories recovered by CAT-Free on the WildTrack test sequence.
  (c) WildTrack MODA plotted against the number of target-scene resources each method uses. MODA provides a common comparison across all listed methods. CAT-Free achieves {89.2 MODA}.
  }
  \label{fig:teaser}
\end{center}%
}]

\begin{abstract}

Multi-camera pedestrian detection and tracking often require camera calibration, position annotations, and target-scene training.
CAT-Free removes all three requirements and uses only synchronized RGB video.
From this video, CAT-Free estimates camera poses and the ground plane.
With the estimated poses and ground plane, 
CAT-Free triangulates detections across cameras and links pedestrian positions into tracks.
However, camera errors and mismatched detections produce false positions.
Two geometric filters remove these false positions: one rejects unstable estimates, and the other rejects positions off the ground.
Both filter thresholds are estimated from the unlabeled video.
Without calibration, annotations, or target-scene training, CAT-Free achieves 89.2, 84.1, and 76.1 MODA on WildTrack, MultiviewX, and GMVD.
CAT-Free is also less affected by camera errors than methods that assume known camera parameters, and it outperforms a domain adaptation method without any adaptation.

\end{abstract}

\section{Introduction}

Multi-camera tracking covers wider areas than a single camera.
When one camera loses a pedestrian, another may maintain the track.
However, existing multi-camera methods often require preparation for each new installation.
Moving a camera often requires recalibration because these methods assume known camera positions and orientations.
Other methods require pedestrian annotations or training on target-scene video.
Annotation requires manually labeling pedestrian positions across frames.
Calibration, annotation, and training increase the cost of adapting to a new site.

Recent multi-view pedestrian detection methods achieve high accuracy on WildTrack and MultiviewX, but often rely on known camera parameters and annotated training data
\cite{hou2020mvdet,hou2021mvdetr,teepe2024earlybird,teepe2024tracktacular}.
UMPD~\cite{liu2024umpd}, DCHM~\cite{ma2025dchm}, and MVUDA~\cite{brorsson2025mvuda} do not use target-scene annotations for training.
However, these methods still require known camera parameters and training on target-scene images.
Thus, avoiding target-scene annotations alone does not remove site-specific preparation.

\begin{figure*}[!tp]
\centering
\includegraphics[width=0.85\linewidth]{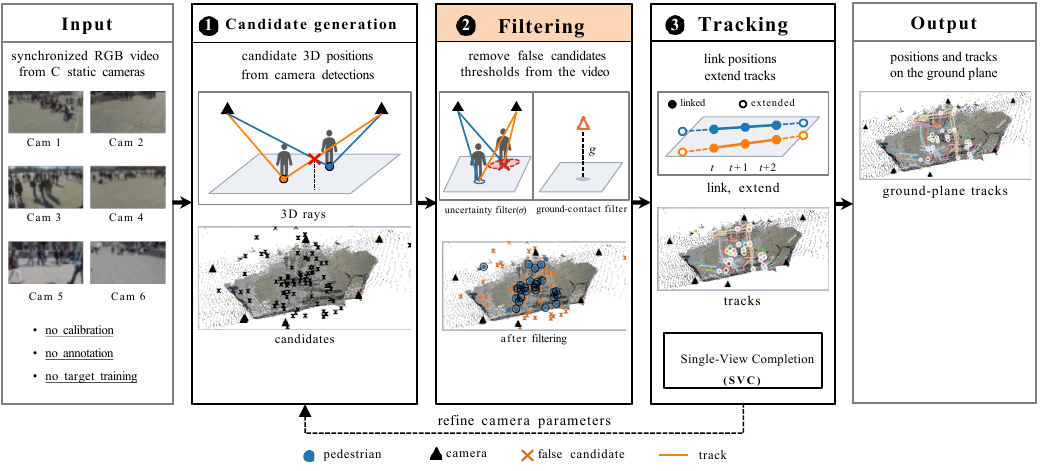}
\caption{CAT-Free pipeline.
CAT-Free takes synchronized RGB video from $C$ static cameras and outputs ground-plane tracks through \cnum{1} candidate generation, \cnum{2} 
false-candidate filtering, and \cnum{3} tracking.
The input shows six of the seven WildTrack cameras.
Black crosses mark generated candidates in \cnum{1}. Orange crosses mark rejected candidates, and blue circles mark retained candidates in \cnum{2}. The orange shading (\cnum{2}) highlights the two adaptive geometric filters.
SVC
in \cnum{3} recovers pedestrians seen by only one camera, and its observations also refine camera estimates (dashed arrow). 
See Secs.~\ref{sec:candidates}, \ref{sec:selection}, and \ref{sec:trajectories} for the three blocks.
}
\label{fig:pipeline}
\end{figure*}

To eliminate per-installation calibration, annotation, and training, we propose CAT-Free, which requires only synchronized RGB video (Fig.~\ref{fig:teaser}(a)).
To recover camera geometry, CAT-Free uses a pretrained 3D reconstruction model to match 3D scene points with image pixels.
Direct camera estimates have larger reprojection errors, so CAT-Free derives camera positions and orientations from the 3D point--pixel correspondences instead (Supplementary Sec.~\ref{sec:vggtcam}).
Cross-camera pedestrian matches further refine these parameters.
Pretrained person detectors find the pedestrians in each view.
With the estimated camera parameters, CAT-Free constructs a 3D ray through the foot point of each detected pedestrian.
CAT-Free triangulates the rays across cameras and takes the point closest to the rays of each pedestrian as the candidate position (Fig.~\ref{fig:pipeline}).
CAT-Free then filters unreliable candidates and links the remaining positions over time (Fig.~\ref{fig:teaser}(b)).
No model is trained or fine-tuned for the new environment, and the same localization and tracking settings are used across datasets, so CAT-Free needs no dataset-specific tuning.

Camera estimation errors can produce false pedestrian positions because each ray depends on the estimated camera parameters.
CAT-Free therefore uses two adaptive geometric filters.
The first filter removes positions that change greatly when the rays shift slightly.
The second filter removes positions where the feet of a pedestrian are far from the ground.
A pedestrian seen by only one camera has no second ray, so triangulation cannot give its 3D position.
Single-view completion (SVC) instead finds the position where the single ray meets the estimated ground, and CAT-Free keeps tracking the pedestrian at that position.
Without known camera parameters, the recovered scene has no fixed metric scale
\cite{hartley2004multiple,wang2025vggt}, so the filters estimate thresholds from input video.

Without calibration or target-scene training, CAT-Free exceeds UMPD and DCHM in MODA on WildTrack and MultiviewX (Table~\ref{tab:published_comparison}).
Both methods require known camera parameters and target-scene training.
Fig.~\ref{fig:teaser}(c) plots the WildTrack MODA of the compared methods against the number of target-scene resources that each method uses.
We further test CAT-Free on GMVD with the same settings.
The results indicate that accurate multi-camera pedestrian detection is possible without preparing camera parameters and without training a model for each new environment.
CAT-Free is also robust to differences between environments.
Methods that assume known camera parameters degrade under camera errors.
CAT-Free also outperforms MVUDA, which adapts to the new environment by training, even though CAT-Free uses no adaptation (Sec.~\ref{sec:crossdata}).

Our contributions are summarized as follows.
\begin{itemize}
    \item
    We present CAT-Free, a multi-camera pedestrian detection and tracking method that uses only RGB video, without known camera parameters, position annotations, or target-scene training.

    \item
    We use two geometric filters to remove unreliable pedestrian positions before the positions are placed on the ground.
    The first filter removes positions that can change greatly with small errors in the rays.
    The second filter removes positions where the feet of a pedestrian appear above or below the ground.

    \item
    Without the calibration, annotations, or target-scene training, 
    CAT-Free achieves 89.2, 84.1, and 76.1 MODA on WildTrack, MultiviewX, and GMVD, approaching methods that use all three on WildTrack (Table~\ref{tab:published_comparison}).

    \item
    CAT-Free is robust to differences between environments: it is less affected by camera errors than methods that assume known camera parameters (Table~\ref{tab:robust}), and it outperforms MVUDA, which adapts to the new environment, without any adaptation (Table~\ref{tab:cross}).

\end{itemize}

\section{Related Work}

MVDet~\cite{hou2020mvdet}, MVDeTr~\cite{hou2021mvdetr}, EarlyBird~\cite{teepe2024earlybird}, and TrackTacular~\cite{teepe2024tracktacular} aggregate features from several cameras on a shared top-down map and achieve high accuracy.
These methods need all three requirements: known camera parameters, position annotations, and training in each new environment.
Some methods reduce these requirements but do not completely remove them.
Probabilistic occupancy maps~\cite{fleuret2008pom} need no training but still require known camera parameters.
UMPD~\cite{liu2024umpd} and DCHM~\cite{ma2025dchm} need no manual annotations, but they require known camera parameters and training on images from the new environment.
MVUDA~\cite{brorsson2025mvuda} adapts a model to a new environment from unlabeled images, but it requires training on images from that environment.

Uncalibrated 3D reconstruction models such as DUSt3R~\cite{wang2024dust3r} and VGGT~\cite{wang2025vggt} estimate camera poses and scene structure from images.
CAT-Free uses VGGT-$\Omega$~\cite{wang2026vggtomega} to derive camera geometry from 3D point--pixel correspondences before ray triangulation (Sec.~\ref{sec:candidates}).


\section{Methodology}
\label{sec:method}

CAT-Free converts RGB video into ground-plane tracks in three blocks (Fig.~\ref{fig:pipeline}).
We model the ground as a plane and represent pedestrian positions in its bird's-eye view (BEV). A position on this plane determines a 3D point, and all views of a pedestrian share this point.
Moreover, an occlusion in one view can be compensated by the other views~\cite{hou2020mvdet,teepe2024tracktacular}.
The input is synchronized RGB video from $C$ static cameras.
VGGT-$\Omega$ first reconstructs a 3D point cloud of the scene, CAT-Free estimates the camera parameters from the point cloud, and a pedestrian detector finds the pedestrians in every image.
The camera parameters are the position and orientation of each camera.
Block \cnum{1} casts a 3D ray from the camera through each detection and intersects the rays of the same pedestrian to generate candidate 3D positions.
Block \cnum{2} removes false candidates with two adaptive geometric filters.
Block \cnum{3} links the remaining positions over time into tracks, keeps tracking the pedestrians that only one camera sees, and outputs the ground-plane tracks of all pedestrians.

Two problems arise.
First, the camera parameters estimated from RGB contain errors, and the reconstructed scene has an unknown scale, so a fixed threshold in metres cannot remove false candidates.
Second, a pedestrian that only one camera sees has only one ray, so block \cnum{1} finds no intersection.
The filters in block \cnum{2} address the first problem, and single-view completion (SVC) in block \cnum{3} addresses the second problem.
Each block has a distinct role:

\begin{enumerate}\setlength{\itemsep}{0pt}\setlength{\topsep}{1pt}\setlength{\parsep}{0pt}

\item \textbf{Ray-based candidate generation} (block \cnum{1}) operates on uncalibrated cameras. CAT-Free casts a 3D ray through the foot point of every detection and takes the intersection of the rays of one pedestrian as its candidate position (Sec.~\ref{sec:candidates}).

\item \textbf{Two adaptive geometric filters} (block \cnum{2}) are the uncertainty filter and the ground-contact filter. Each filter estimates its threshold automatically from the input video (Sec.~\ref{sec:selection}).

\item \textbf{Single-view completion (SVC)} (block \cnum{3}) keeps tracking pedestrians that only one camera sees (Sec.~\ref{sec:trajectories}).
\end{enumerate}

\subsection{Candidate generation (\cnum{1})}
\label{sec:candidates}
\label{sec:camera_config}
\label{sec:rays}
\label{sec:matching}
\label{sec:position_estimation}

\begin{figure*}[t]
\centering
\includegraphics[width=0.85\linewidth]{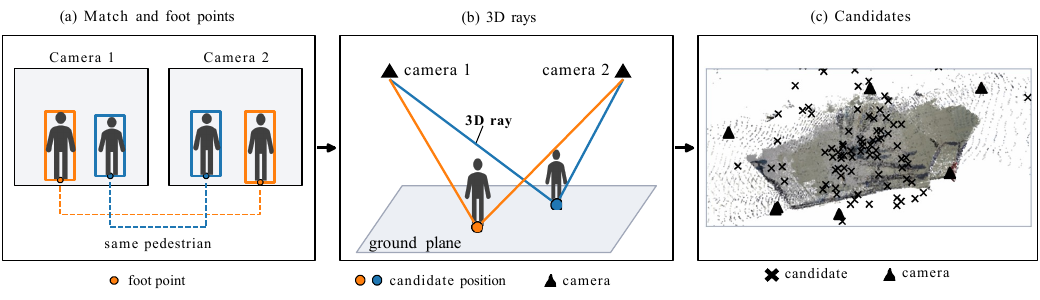}
\caption{The details of candidate generation in block \cnum{1} of Fig.~\ref{fig:pipeline}. Panels (a) and (b) are schematic. (a) The same pedestrian is matched across cameras, and the foot point of each box gives a 3D ray. (b) The rays of one pedestrian meet at a candidate position. (c) An example of all candidates on WildTrack frame 1900.}
\label{fig:candidates}
\end{figure*}

Block \cnum{1} locates pedestrians in 3D from RGB video alone.
First, we detect pedestrians in each image and take the foot point of each pedestrian from the bounding box.
The foot point gives the direction in which the pedestrian stands, but not the distance from the camera.
We therefore use the estimated camera parameters and extend a 3D ray from the camera center toward the foot point.
The actual feet of the pedestrian lie somewhere on this ray (Fig.~\ref{fig:candidates}(a)).
When another camera sees the same pedestrian, the rays of the two cameras intersect near the feet (Fig.~\ref{fig:candidates}(b)).
We use this property to estimate the 3D position of the pedestrian as the point closest to the rays.
This operation is called triangulation.
However, the rays of different pedestrians can also come close by chance.
We therefore do not fix the position at this stage and treat it as a candidate position.
The filters of block \cnum{2} then select the correct candidates (Sec.~\ref{sec:selection}).

\paragraph{Camera geometry estimation.}

CAT-Free estimates camera positions and orientations before casting the rays in Fig.~\ref{fig:candidates}.
VGGT-$\Omega$~\cite{wang2026vggtomega} reconstructs a 3D point cloud and provides 3D point--pixel correspondences.
CAT-Free derives camera parameters from these correspondences rather than using VGGT-$\Omega$'s direct camera estimates.
The derived parameters have smaller reprojection errors than VGGT-$\Omega$ on all seven WildTrack cameras (Supplementary Sec.~\ref{sec:vggtcam}).

\paragraph{Pedestrian detection.}

YOLO11x~\cite{jocher2024yolo11} detects pedestrians in each camera independently.
For each pedestrian, YOLO11x outputs a bounding box and pose keypoints, which are the image positions of body joints such as the ankles.
We use the publicly available YOLO11x model trained on COCO.
The model generalizes across scenes, so training on the target scene is not needed.
OSNet~\cite{zhou2019osnet} extracts an appearance feature from each detection.
The appearance feature is a vector that describes how the pedestrian looks.
We use the appearance feature for three purposes: to match pedestrians across cameras, to link positions over time, and to join tracklets (Sec.~\ref{sec:trajectories}).
Two tracklets are joined when the cosine similarity of their OSNet features exceeds a threshold.

\paragraph{3D rays and their reliability.}

The foot point is where the pedestrian touches the ground (Fig.~\ref{fig:candidates}(a)), so the rays of the same pedestrian meet at the foot point whatever the body height.
We take the foot point from the ankle keypoint when the pose estimator confirms the ankle, and from the bottom center of the bounding box otherwise.
The ray does not need a visible ankle.
For detection $i$, the camera center is $\mathbf{o}_i$, and the unit ray direction is $\mathbf{d}_i$.
The symbol $\hat{\mathbf{x}}$ denotes the estimated 3D position of a pedestrian, which is the point closest to the rays of that pedestrian.

We measure the error of a ray as an angle, because an angle does not depend on the unknown scale of the reconstruction.
The angular error of a single ray has a standard deviation $\sigma_\theta$, which we estimate from the input video (Supplementary Sec.~\ref{sec:sigmatheta}).
A ray through a confirmed ankle is more accurate than a ray through the bottom center of the box, and a pedestrian far from the camera has a less precise foot point.
We give each ray $i$ a reliability weight $w_i \in (0,1]$ that is larger for a more accurate ray.

\paragraph{Cross-camera matching and triangulation.}

Before triangulation, we must decide which detections in different cameras show the same pedestrian, because only the rays of one pedestrian meet.
We start with each detection as its own group and merge two groups at a time in the order of the merge cost.
The merge cost is the normalized ray distance of the merged group minus cosine similarity of OSNet features.
Groups whose rays nearly meet and whose appearances are similar merge first.
Two groups are merged only if no camera has a detection in both groups, their appearances are similar, and their rays meet at one point.
We keep the triangulated position $\hat{\mathbf{x}}$ in 3D, so block~\cnum{2} can reject wrong matches by the height above the ground plane (Sec.~\ref{sec:selection}).
Block~\cnum{1} thus gives 3D candidate positions of pedestrians from RGB video without a depth sensor, known camera parameters, or training on the target scene (Fig.~\ref{fig:candidates}(c)).

\subsection{Filtering (\cnum{2})}
\label{sec:selection}

\begin{figure*}[t]
\centering
\includegraphics[width=0.9\linewidth]{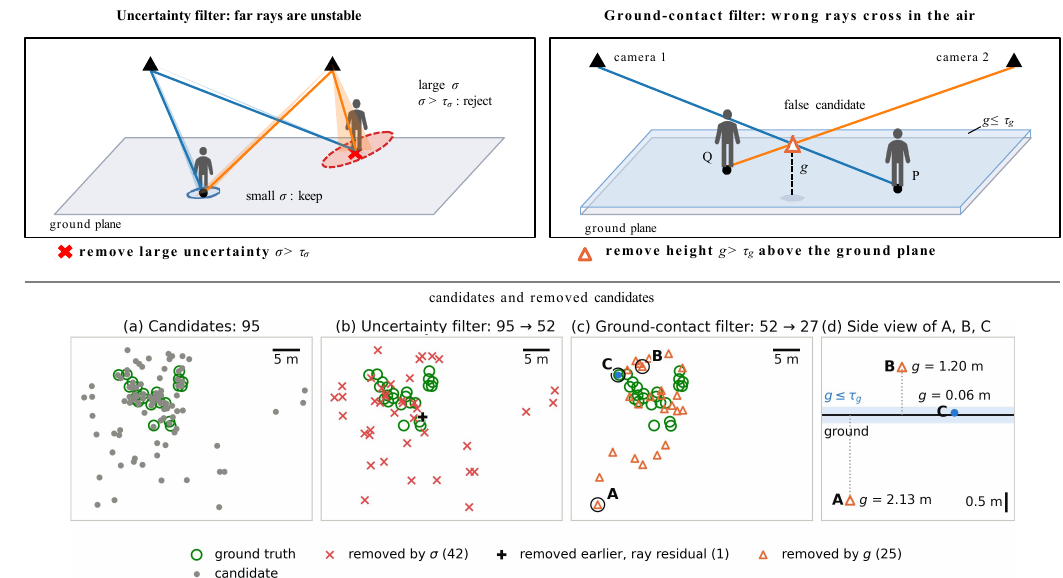}
\caption{
False candidates and their removal by two geometric filters. Top left: Shaded wedges show angular errors, and the ellipse shows uncertainty $\sigma$ (Eq.~\eqref{eq:sigma}). The uncertainty filter keeps the left pedestrian but rejects the distant one with $\sigma > \tau_\sigma$. Top right: Rays from pedestrians P and Q intersect above the ground, creating a false candidate. The ground-contact filter rejects it when $g > \tau_g$ (Eq.~\eqref{eq:height}); the blue band marks $g \leq \tau_g$. Bottom: WildTrack frame 1900. (a) All candidates. (b) The uncertainty filter removes 42 candidates (red crosses), and the ray-residual check removes one (black plus). (c) The ground-contact filter removes 25 candidates (orange triangles). (d) Side view of rejected candidates A and B and retained candidate C. Ground truth and scale bars serve only as visual references.
}
\label{fig:schematic}
\end{figure*}

Block \cnum{1} generates two kinds of false candidates.
First, some candidates have imprecise positions.
A small angular error of each ray moves the intersection far, so a pedestrian far from one of the cameras that see it is localized only roughly.
Second, some candidates match different pedestrians.
The rays of two different pedestrians can intersect in the air.
The uncertainty filter removes the first kind, and the ground-contact filter removes the second kind (block~\cnum{2} in Fig.~\ref{fig:pipeline}).
Fig.~\ref{fig:schematic} shows one example of each.
In the top left, a pedestrian far from camera 1 is localized only roughly.
In the top right, the rays of two different pedestrians cross above the ground.
The bottom row shows the effect of the two filters on WildTrack frame 1900.
On this frame, the 95 candidates shrink to 52 after the uncertainty filter and to 27 after the ground-contact filter, for 23 annotated pedestrians (Fig.~\ref{fig:schematic}, bottom, (a) to (c)).
CAT-Free estimates both thresholds from the input video, so the filters need no metric scale, no position annotations, and no tuning for each scene.
Thus, neither filter needs site-specific calibration, annotations, or tuning.

\paragraph{Uncertainty filter.}

The uncertainty filter keeps only the candidates whose positions the cameras fix precisely.
A candidate is poorly constrained when a camera that sees it is far away, because a small angular error of the ray moves the intersection far (Fig.~\ref{fig:schematic}, top left).
We measure how well a candidate is constrained by the uncertainty $\sigma$.
The uncertainty $\sigma$ is the spread of the candidate position when every ray carries the angular error $\sigma_\theta$.

Ray $i$ constrains $\hat{\mathbf{x}}$ only in the directions perpendicular to the ray direction $\mathbf{d}_i$, which is the arrow from camera $i$ to the pedestrian.
The standard deviation of this constraint is about $\sigma_\theta L_i$, where $L_i=\|\hat{\mathbf{x}}-\mathbf{o}_i\|$ is the distance from camera $i$ to the candidate.
Combining these rays with their reliability weights $w_i$ (Sec.~\ref{sec:rays}) gives the covariance in Eq.~\eqref{eq:sigma},
\begin{equation}
\label{eq:sigma}
\Sigma_{\hat{\mathbf{x}}}
=
\left[
\sum_{i\in\mathcal{R}}
\frac{w_i}{(\sigma_\theta L_i)^2}
\left(\mathbf{I}-\mathbf{d}_i\mathbf{d}_i^\top\right)
\right]^{-1},
\end{equation}

where $\mathcal{R}$ denotes the rays supporting $\hat{\mathbf{x}}$.
The localization uncertainty $\sigma$ is the root-mean-square standard deviation of $\Sigma_{\hat{\mathbf{x}}}$ in Eq.~\eqref{eq:sigma} along the two directions of the estimated ground plane.
The uncertainty $\sigma$ describes the candidate position and differs from the angular error $\sigma_\theta$ of a ray.
We accept a candidate when
\begin{equation}
\label{eq:sigmafilter}
\sigma \leq \tau_\sigma,
\qquad
\tau_\sigma =
\exp\left(\mathrm{Otsu}(\{\log \sigma\})\right).
\end{equation}
The reconstruction has an unknown unit, so $\sigma$
also has an unknown unit, and a fixed threshold would not transfer between scenes.
We estimate $\tau_\sigma$ from the input video with Otsu's method~\cite{otsu1979}.
It splits a set of values into two groups without labels.
We apply Otsu's method to $\log\sigma$ of all candidates in the sequence, because $\sigma$ has a strongly skewed distribution with a small number of large values.

\begin{table*}[t]
\centering
\caption{
Comparison of MODA, MODP, MOTA, and IDF1 on WildTrack and MultiviewX, including the use of camera calibration, position annotations, and target-scene training. Methods are grouped by their requirements and further divided into detection and tracking methods. ``--'' denotes unreported results, and bold indicates that CAT-Free requires none of the three resources. }
\label{tab:published_comparison}
\setlength{\tabcolsep}{1.5pt}
\renewcommand{\arraystretch}{1.0}
\scriptsize
\begin{tabular}{@{}ll l ccc rrrr rrrr@{}}
\toprule
& & & & & &
\multicolumn{4}{c}{WildTrack} &
\multicolumn{4}{c}{MultiviewX} \\
\cmidrule(lr){7-10}\cmidrule(lr){11-14}
& & Method &
\shortstack{Known camera\\parameters} &
\shortstack{Target\\annotations} &
\shortstack{Target\\training} &
MODA & MODP & MOTA & IDF1 &
MODA & MODP & MOTA & IDF1 \\
\midrule
\rowcolor{oursblue}\cellcolor{white}\multirow{4}{*}{\scriptsize {\shortstack[l]{Fewer than\\three needed}}} & \cellcolor{white} & CAT-Free (ours)
& \textbf{No} & \textbf{No} & \textbf{No}
& 89.2 & 73.0 & 90.8 & 90.4
& 84.1 & 73.1 & 82.7 & 81.6 \\
 &  & UMPD~\cite{liu2024umpd}
& Yes & No & Yes
& 76.6 & 61.2 & -- & --
& 67.5 & 79.4 & -- & -- \\
 &  & DCHM~\cite{ma2025dchm}
& Yes & No & Yes
& 84.2 & 80.3 & -- & --
& 78.4 & 82.3 & -- & -- \\
&  & MVUDA~\cite{brorsson2025mvuda}
& Yes & No & Yes
& 85.4 & 75.3 & -- & --
& 82.2 & 75.6 & -- & -- \\
\midrule
\multirow{8}{*}{\scriptsize {\shortstack[l]{All three\\needed}}} & \multirow{3}{*}{\scriptsize {Detection}} & MVDet~\cite{hou2020mvdet}
& Yes & Yes & Yes
& 88.2 & 75.7 & -- & --
& 83.9 & 79.6 & -- & -- \\
 &  & MVDeTr~\cite{hou2021mvdetr}
& Yes & Yes & Yes
& 91.5 & 82.1 & -- & --
& 93.7 & 91.3 & -- & -- \\
 &  & MV2GF~\cite{yamane2026mv2gf}
& Yes & Yes & Yes
& 94.7 & 82.9 & -- & --
& -- & -- & -- & -- \\
\cmidrule(l){2-14}
 & \multirow{5}{*}{\scriptsize {Tracking}} & EarlyBird~\cite{teepe2024earlybird}
& Yes & Yes & Yes
& 91.2 & 81.8 & 89.5 & 92.3
& 94.2 & 90.1 & 88.4 & 82.4 \\
 &  & TrackTacular~\cite{teepe2024tracktacular}
& Yes & Yes & Yes
& 93.2 & 77.5 & 90.2 & 93.6
& 96.5 & 75.0 & 92.4 & 85.6 \\
 &  & ATLASFusion~\cite{toida2025atlasfusion}
& Yes & Yes & Yes
& 93.6 & 82.1 & 92.4 & 95.9
& 96.5 & 89.2 & 92.5 & 85.0 \\
 &  & {Beyond Centers~\cite{ali2026beyondcenters}}
& Yes & Yes & Yes
& 91.0 & 82.6 & 89.6 & 92.5
& -- & -- & -- & -- \\
 &  & {MVTrackTrans~\cite{zhang2026mvtracktrans}}
& Yes & Yes & Yes
& -- & -- & 91.2 & 94.1
& -- & -- & 89.8 & 72.1 \\
\bottomrule
\end{tabular}
\end{table*}

\paragraph{Ground-contact filter.}

The ground-contact filter removes the wrong matches whose rays intersect precisely but away from the ground.
The feet of a correct pedestrian touch the ground, so a correct candidate lies near the estimated ground plane.
The ground-contact filter rejects candidates whose height above the plane exceeds a threshold (Fig.~\ref{fig:schematic}).

We measure the height $g$ of the estimated 3D position $\hat{\mathbf{x}}$ as its distance to the ground plane $\mathbf{n}^\top\mathbf{x}+b=0$, where $\mathbf{n}$ is the unit normal and $b$ is the offset (Supplementary Sec.~\ref{sec:cameradetail}), 
\begin{equation}
\label{eq:height}
g = |\mathbf{n}^\top \hat{\mathbf{x}} + b|.
\end{equation}

We accept a candidate when
\begin{equation}
\label{eq:gfilter}
g \leq \tau_g,
\qquad
\tau_g =
\exp\left(\mathrm{Otsu}(\{\log g\})\right).
\end{equation}
Like $\sigma$, the height $g$ in Eq.~\eqref{eq:gfilter} has the unknown unit of the reconstruction.
We estimate $\tau_g$ from the input video with Otsu's method.
We apply Otsu's method to $\log g$ of the candidates that pass the uncertainty filter.
The two filters thus work in this order: the uncertainty filter first removes uncertain positions, and the ground-contact filter then tests the height above the ground.
Supplementary Sec.~\ref{sec:selectdetail} gives further details of the candidate selection.

\subsection{Tracking (\cnum{3})}
\label{sec:trajectories}

Block \cnum{3} turns the positions accepted by block \cnum{2} into tracks (Fig.~\ref{fig:pipeline}).
A pedestrian moves only a short distance between frames, so we link each position to the positions of the next frame within a linking radius.
Two pedestrians who walk side by side can both lie within this radius.
We define the link cost as the ground-plane distance plus the OSNet appearance distance.
The Hungarian algorithm then selects the one-to-one links with the lowest total cost.

\begin{table}[t]
\centering
\caption{
Cross-dataset evaluation. Source only, GMVD, and MVUDA use source-dataset training, while CAT-Free requires no training. MVUDA uses unlabeled target images for adaptation. Results for the other methods come from their original papers.
}
\label{tab:cross}
\footnotesize
\setlength{\tabcolsep}{3pt}
\begin{tabular}{@{}l l rrrr@{}}
\toprule
Setting & Method & MODA & MODP & MOTA & IDF1 \\
\midrule
\multirow{3}{*}{\shortstack[l]{WildTrack\\$\to$\\MultiviewX}} & Source only~\cite{brorsson2025mvuda} & 35.7 & 66.5 & -- & -- \\
 & MVUDA~\cite{brorsson2025mvuda} & 82.2 & \textbf{75.6} & -- & -- \\
 & \cellcolor{oursblue}CAT-Free (ours) & \cellcolor{oursblue}\textbf{84.1} & \cellcolor{oursblue}73.1 & \cellcolor{oursblue}82.7 & \cellcolor{oursblue}81.6 \\
\midrule
\multirow{4}{*}{\shortstack[l]{MultiviewX\\$\to$\\WildTrack}} & Source only~\cite{brorsson2025mvuda} & 70.0 & 73.6 & -- & -- \\
 & GMVD~\cite{vora2023gmvd} & 70.7 & 73.8 & -- & -- \\
 & MVUDA~\cite{brorsson2025mvuda} & 85.4 & \textbf{75.3} & -- & -- \\
 & \cellcolor{oursblue}CAT-Free (ours) & \cellcolor{oursblue}\textbf{89.2} & \cellcolor{oursblue}73.0 & \cellcolor{oursblue}90.8 & \cellcolor{oursblue}90.4 \\
\bottomrule
\end{tabular}
\end{table}

\begin{table}[t]
\centering
\caption{
Robustness to camera parameter errors.
"No noise" indicates ground-truth parameters.
"Noise" adds Gaussian noise to rotation and position, using the mean and standard deviation of CAT-Free's estimation errors over the cameras (degrees and cm).
"Estimated" uses CAT-Free's camera estimates.
CAT-Free requires no known camera parameters.
}
\label{tab:robust}
\scriptsize
\setlength{\tabcolsep}{2.2pt}
\begin{tabular}{@{}c l rrrr@{}}
\toprule
Data & Cameras & MODA & MODP & MOTA & IDF1 \\
\midrule
\multirow{13}{*}{MultiviewX} & \multicolumn{5}{@{}l}{\textit{TrackTacular}} \\
 & No noise & 95.9 & 89.6 & 90.8 & 79.9 \\
 & Noise ($1.31^\circ$, 5.9\,cm) & 24.1 & 53.1 & 37.4 & 46.4 \\
 & Estimated & 94.5 & 83.4 & 89.5 & 80.2 \\
\cmidrule(l){2-6}
 & \multicolumn{5}{@{}l}{\textit{ATLASFusion}} \\
 & No noise & 95.9 & 88.8 & 91.9 & 82.9 \\
 & Noise ($1.31^\circ$, 5.9\,cm) & 16.1 & 51.7 & 27.3 & 44.7 \\
 & Estimated & 94.4 & 81.9 & 90.5 & 83.2 \\
\cmidrule(l){2-6}
 & \multicolumn{5}{@{}l}{\textit{MVTrackTrans}} \\
 & No noise & 92.7 & 86.3 & 88.8 & 66.3 \\
 & Noise ($1.31^\circ$, 5.9\,cm) & 11.2 & 58.9 & 13.6 & 18.4 \\
 & Estimated & 88.7 & 80.5 & 86.0 & 66.8 \\
\cmidrule(l){2-6}
 & \cellcolor{oursblue}CAT-Free (ours) & \cellcolor{oursblue}84.1 & \cellcolor{oursblue}73.1 & \cellcolor{oursblue}82.7 & \cellcolor{oursblue}81.6 \\
\midrule
\multirow{9}{*}{WildTrack} & \multicolumn{5}{@{}l}{\textit{TrackTacular}} \\
 & No noise & 90.7 & 81.0 & 91.1 & 94.9 \\
 & Noise ($3.75^\circ$, 30.4\,cm) & 0.8 & 41.3 & 19.7 & 36.6 \\
 & Estimated & 37.9 & 52.6 & 57.0 & 64.1 \\
\cmidrule(l){2-6}
 & \multicolumn{5}{@{}l}{\textit{ATLASFusion}} \\
 & No noise & 90.1 & 79.3 & 92.2 & 95.4 \\
 & Noise ($3.75^\circ$, 30.4\,cm) & 1.2 & 41.8 & 17.1 & 40.0 \\
 & Estimated & 30.1 & 38.3 & 56.0 & 71.7 \\
\cmidrule(l){2-6}
 & \cellcolor{oursblue}CAT-Free (ours) & \cellcolor{oursblue}89.2 & \cellcolor{oursblue}73.0 & \cellcolor{oursblue}90.8 & \cellcolor{oursblue}90.4 \\
\bottomrule
\end{tabular}
\end{table}

\begin{table*}[t]
\centering
\caption{
Ablation of the three components, added sequentially (\checkmark indicates use). Row (a) uses only \cnum{1}, with camera support replacing the filters. Component \cnum{1} is essential for generating candidates. $^\dagger$Candidate refinement is re-triangulation, which discards outlier rays and triangulates again, and a score-based recheck of the candidates. Row (e) is CAT-Free. The best value in each column is in bold.
}
\label{tab:ablation}
\footnotesize
\setlength{\tabcolsep}{1.8pt}
\begin{tabular}{@{}l ccccc rrrr rrrr@{}}
\toprule
& \shortstack{\cnum{1}\\candidates\\(3.1)} & \shortstack{\cnum{2}\\uncertainty\\filter} & \shortstack{\cnum{2}\\ground-contact\\filter} & \shortstack{candidate\\refinement$^\dagger$} & \shortstack{\cnum{3}\\SVC\\(3.3)} & \multicolumn{4}{c}{MODA} & \multicolumn{4}{c}{IDF1} \\
\cmidrule(lr){7-10}\cmidrule(lr){11-14}
& & & & & & WildTrack & MultiviewX & GMVD & Mean & WildTrack & MultiviewX & GMVD & Mean \\
\midrule
(a) & \checkmark &  &  &  &  & 78.9 & 73.0 & 67.9 & 73.3 & 56.3 & 55.5 & 59.6 & {57.1} \\
(b) & \checkmark & \checkmark &  &  &  & 76.5 & 73.4 & 71.9 & 73.9 & 58.9 & 56.0 & 60.3 & {58.4} \\
(c) & \checkmark & \checkmark & \checkmark &  &  & 80.3 & 75.0 & 65.3 & {73.5} & 60.0 & 57.8 & 53.9 & {57.2} \\
(d) & \checkmark & \checkmark & \checkmark & \checkmark &  & 87.5 & 82.8 & 75.2 & {81.8} & {71.9} & {64.2} & {61.0} & {65.7} \\
(e) & \checkmark & \checkmark & \checkmark & \checkmark & \checkmark & \textbf{89.2} & \textbf{84.1} & \textbf{76.1} & \textbf{83.1} & {\textbf{90.4}} & {\textbf{81.6}} & {\textbf{82.1}} & {\textbf{84.7}} \\
\bottomrule
\end{tabular}
\end{table*}

\paragraph{Single-view completion (SVC).}
\label{sec:singleview}

Block \cnum{1} needs rays from at least two different cameras to locate a pedestrian in 3D, so block \cnum{1} cannot locate a pedestrian that only one camera sees.
These missed pedestrians reduce detection and tracking accuracy.
SVC addresses the problem in two steps and locates the 3D position of each pedestrian that only one camera sees.
First, SVC works in each camera independently and compares two tracklets that appear at different times.
A tracklet is a short track of the detections of one camera.
If the cosine similarity between the OSNet features of the two tracklets exceeds a threshold, SVC regards them as the same pedestrian and joins them into one ID.
Second, SVC intersects the ray through the foot point in the image with the estimated ground plane.
A single ray alone does not give the distance to the pedestrian, but the feet of the pedestrian lie on the ground.
We take the point where the ray, with origin $\mathbf{o}$ and direction $\mathbf{d}$, meets the ground plane as the 3D position (Eq.~\eqref{eq:svc}),
\begin{equation}
\label{eq:svc}
\mathbf{x}=\mathbf{o}+s\,\mathbf{d},
\qquad
s=-\frac{\mathbf{n}^\top\mathbf{o}+b}{\mathbf{n}^\top\mathbf{d}}.
\end{equation}
where $s$ is the distance from the camera to the point where the ray meets the ground plane $\mathbf{n}^\top\mathbf{x}+b=0$.

SVC also improves the camera estimation.
We add the pedestrians placed on the ground, as 3D reference points, to the refinement of the camera parameters.
The dashed arrow from SVC in block~\cnum{3} back to block~\cnum{1} in Fig.~\ref{fig:pipeline} shows this feedback (Supplementary Sec.~\ref{sec:cameradetail}).
SVC thus locates pedestrians visible to only one camera, including those behind obstacles or at the edge of the covered area.
CAT-Free also keeps the same ID for a pedestrian who disappears temporarily (Sec.~\ref{sec:ablation}).

\subsection{Candidate refinement}
\label{sec:refinement}

Rays from wrongly matched detections can remain even after the matching check (Sec.~\ref{sec:position_estimation}) and degrade the triangulation in block~\cnum{1} and the filtering in block~\cnum{2} (Sec.~\ref{sec:selection}).
We remove rays far from the estimated point and triangulate again from the remaining rays, and recheck the candidates after the filters with a score computed from seven image-derived cues (Supplementary Sec.~\ref{sec:refinementdetail}).
Sec.~\ref{sec:ablation} evaluates the effect.

\section{Experiments}
\label{sec:results}

\begin{figure*}[t]
\centering
\includegraphics[width=0.8\textwidth,trim=0 38 0 100,clip]{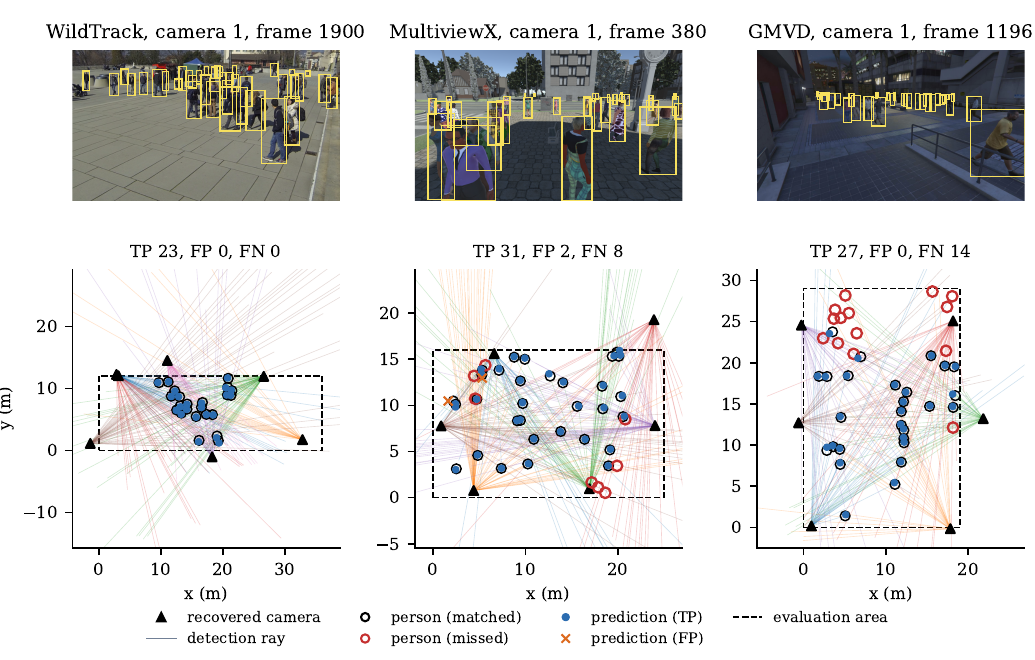}
\caption{Qualitative results in BEV space on WildTrack, MultiviewX, and GMVD.
Blue dots are correct predictions, orange crosses are false positives, red circles are missed pedestrians, black triangles are the recovered cameras, and the dashed box is the evaluation area.
}
\label{fig:qualitative}
\end{figure*}

\paragraph{Setup.}

We evaluate on WildTrack~\cite{chavdarova2018wildtrack} (7 cameras), MultiviewX~\cite{hou2020mvdet} (6 cameras), and GMVD~\cite{vora2023gmvd} (7 scenes).
We use the test sets of all three datasets.

CAT-Free uses synchronized RGB video and frozen pretrained models.
It estimates $\sigma_\theta$ and two filter thresholds from each video. For evaluation only, we apply a 7-DoF Umeyama alignment~\cite{umeyama1991} using estimated and ground-truth camera centers.
CAT-Free uses neither the alignment nor the ground-truth camera centers.

Following prior work~\cite{chavdarova2018wildtrack,teepe2024tracktacular}, MODA and MODP use a 0.5\,m threshold, while MOTA, MOTP, and IDF1 use 1\,m.
MODA~\cite{kasturi2009moda} measures detection errors, and MODP measures localization accuracy.
MOTA~\cite{bernardin2008clear} also counts identity switches, while IDF1~\cite{ristani2016idf1} measures identity consistency.
We use annotations only to compute metrics after inference.
Supplementary Sec.~\ref{sec:expdetails} shows the Setup details.

\subsection{Main results}
\label{sec:mainresults}

\paragraph{CAT-Free works without per-installation calibration, annotation, or training.}

For a fair comparison, we compare CAT-Free against published methods that also output pedestrian positions in BEV.
Table~\ref{tab:published_comparison} lists the requirements of each method, which are known camera parameters, target annotations, and target training.
Each requirement is a separate preparation task.
The table also reports detection accuracy (MODA, MODP) and tracking accuracy (MOTA, IDF1).
All methods except MVTrackTrans report detection accuracy.
Only five methods
report tracking accuracy, so Table~\ref{tab:published_comparison} marks the other methods with ``--''.

Table~\ref{tab:published_comparison} shows that CAT-Free outperforms UMPD and DCHM in MODA by 5.0 to 16.6 percentage points on WildTrack and MultiviewX (Fig.~\ref{fig:teaser}(c)).
UMPD and DCHM avoid only the annotations, whereas CAT-Free avoids all three requirements.
These comparisons show that CAT-Free attains higher MODA despite requiring fewer target-scene resources.
CAT-Free outputs ground-plane positions and tracks from synchronized RGB video alone.

CAT-Free instead targets new or changing camera installations where calibration, annotation, and target-scene training are impractical.
In such installations, we need only synchronized RGB video to use CAT-Free.

Five methods report tracking accuracy: EarlyBird, TrackTacular, ATLASFusion, Beyond Centers, and MVTrackTrans.
All five methods need all three requirements.
On the real-video WildTrack dataset, CAT-Free achieves 90.8 MOTA, exceeding EarlyBird, Beyond Centers, and TrackTacular and trailing the strongest 
baseline by 1.6 points.
CAT-Free uses YOLO11x pretrained for person detection on real images, whereas the compared methods are trained on the target videos from ImageNet-pretrained backbones.
This result shows competitive tracking accuracy without target-scene training.
On MultiviewX, CAT-Free achieves 81.6 IDF1, outperforming MVTrackTrans and coming within 0.8 points of EarlyBird.

\paragraph{Qualitative results.}

Fig.~\ref{fig:qualitative} visualizes the middle frame of the evaluation sequence of each dataset.
On WildTrack, CAT-Free finds all 23 annotated pedestrians with no false positive, although many pedestrians are close to each other.
On MultiviewX, CAT-Free finds 31 of the 39 annotated pedestrians with 2 false positives.
On GMVD, CAT-Free finds 27 of the 41 annotated pedestrians with no false positive.
Over the full evaluation sets, CAT-Free achieves a precision of 98.1\%, 97.7\%, and 97.3\% on WildTrack, MultiviewX, and GMVD, averaged over the ten videos for GMVD.
This indicates that false positives are rare on all three datasets.
The precision is comparable to that of methods trained on the target dataset~\cite{teepe2024tracktacular}.

\subsection{Cross-dataset evaluation}
\label{sec:crossdata}

Table~\ref{tab:cross} compares CAT-Free with methods that transfer from one dataset to another.
Source only, GMVD, and MVUDA use source-dataset training; CAT-Free needs no training on either evaluation dataset.
We focus on MVUDA because it has the highest MODA among the other methods and is the only one that uses the target images.
CAT-Free outperforms MVUDA, which adapts to the unlabeled target images, by 1.9 percentage points on MultiviewX and by 3.8 percentage points on WildTrack.
The source-only models underperform CAT-Free on the target dataset.
CAT-Free avoids source-to-target adaptation because neither evaluation dataset is used for training.

\subsection{Robustness to camera parameter errors}
\label{sec:robust}

We test how camera parameter errors affect methods that assume known calibration.
We compare CAT-Free with baselines under noisy camera parameters on MultiviewX (TrackTacular, ATLASFusion, MVTrackTrans) and WildTrack (TrackTacular, ATLASFusion) in Table~\ref{tab:robust}.
We also evaluate the same methods with the camera parameters that CAT-Free estimates from the video.
The rotation noise is a Gaussian noise per axis, in degrees, and we multiply $\mathbf{R}$ from the left by it.
The position noise is a Gaussian noise per axis, in centimeters, and we add it to $\mathbf{t}$.
The mean and standard deviation of each Gaussian are those of the errors of the camera parameters that CAT-Free estimates over the cameras.
The means and standard deviations are $3.75\pm1.49^\circ$ and $30.4\pm6.0$\,cm on WildTrack and $1.31\pm2.26^\circ$ and $5.9\pm2.2$\,cm on MultiviewX.
Each value is the average of five runs.

With CAT-Free-level camera errors, CAT-Free exceeds the calibrated baselines on WildTrack in all 
metrics and MVTrackTrans on MultiviewX in MODA, MOTA, and IDF1 (Table~\ref{tab:robust}).
The advantage is largest on WildTrack, where the camera error is larger.
The IDF1 of the other methods drops less than their MODA, which suggests that the camera error mainly shifts the positions of the pedestrians.

With the camera parameters that CAT-Free estimates, the other methods retain most of their accuracy on MultiviewX, where the estimated camera parameters have smaller errors.
On WildTrack, the estimated camera parameters have larger errors.
The cause may be lens distortion: the real cameras of WildTrack exhibit lens distortion, whereas the synthetic cameras of MultiviewX exhibit little.
CAT-Free achieves 89.2 MODA, more than twice the MODA of TrackTacular and ATLASFusion.
This suggests that modeling the angular error of the rays (Secs.~\ref{sec:candidates} and~\ref{sec:selection}) is useful when the camera parameters are estimated.

\subsection{Ablation study}
\label{sec:ablation}

\paragraph{Each component addresses a different error.}

Table~\ref{tab:ablation} tests each component by adding it sequentially from row (a) to full CAT-Free in row (e).
Candidate refinement improves MODA the most, and SVC improves IDF1 the most, so SVC mainly keeps the identities of the pedestrians.
The two filters change the mean MODA only slightly, because the ground-contact filter improves MODA on WildTrack and MultiviewX and degrades it on GMVD.
The accuracy drop on GMVD comes from two of the ten videos, in which the threshold of the ground-contact filter becomes too small and removes more than half of the annotated pedestrians (Supplementary Sec.~\ref{sec:analysis}).
Together, the filters, candidate refinement, and SVC raise MODA and IDF1 across all three datasets, so they complement each other.

\section{Conclusion}

Many multi-camera detection and tracking methods require calibration, position annotations, or target-scene training.
CAT-Free removes all three requirements using only synchronized RGB video.
Two adaptive filters reject false positions, while SVC tracks pedestrians visible to only one camera. CAT-Free achieves competitive accuracy without target-scene preparation.


{
    \small
    \bibliographystyle{ieeenat_fullname}
    \bibliography{main}
}

\clearpage
\appendix
\setcounter{table}{0}\setcounter{figure}{0}
\renewcommand{\thetable}{S\arabic{table}}\renewcommand{\thefigure}{S\arabic{figure}}
\renewcommand{\theHtable}{S\arabic{table}}\renewcommand{\theHfigure}{S\arabic{figure}}
\section{Method Details}
\label{sec:methoddetails}

This section collects every detail of the basic method (Sec.~\ref{sec:method}) that is omitted from the main text so that its main line stays visible: camera parameters, bundle adjustment, ground plane, ray reliability, matching, robust triangulation, candidate selection, and trajectory linking. {Secs.~\ref{sec:refinementdetail} and~\ref{sec:hyperparams} describe the candidate refinement and list the hyperparameters.}

\subsection{Camera parameters, bundle adjustment, and ground plane}
\label{sec:cameradetail}

\paragraph{Camera parameters.}

VGGT-$\Omega$ reconstructs a 3D point cloud of the scene and relates each 3D point to the pixels of each camera image, and we compute the camera parameters (position and orientation) of each camera from these correspondences. {We solve the Perspective-n-Point (PnP) problem, which estimates the position and orientation of a camera from 3D points and their pixels in the image.} {We do not use the camera parameters that VGGT-$\Omega$ outputs directly, because their reprojection error is larger than that of the back-computed parameters (Sec.~\ref{sec:vggtcam}).} 
We use the same checkpoint for all datasets, and we combine the per-frame results into a fixed multi-camera rig.

\paragraph{Bundle adjustment.}

We refine the camera parameters by bundle adjustment with a Huber loss, using pedestrians matched across cameras and supported by at least three cameras as 3D reference points. 
We repeat matching and refinement for a fixed number of iterations and keep the result with the smallest median geometric error. {In addition, SVC places pedestrians seen by only one camera on the ground, and we add these pedestrians to the 3D reference points as well (Sec.~\ref{sec:trajectories}).}

\paragraph{Ground plane.}

We fit the plane $\mathbf{n}^\top\mathbf{x}+b=0$ to the reconstructed point cloud by RANSAC, where $\mathbf{n}$ is its unit normal and $b$ its offset, and take this plane as the ground. We reject planes with too few inliers, planes with inconsistent camera sides, and planes with an implausible spread of camera heights (wall-like planes).

\subsection{Ray construction}
\label{sec:raydetail}

Each detection defines a 3D ray from the camera toward the detected pedestrian.
The ray starts at the camera center and passes through the foot point of the detection.
We take the foot point from the left and right ankle keypoints of the pose estimator, which follow the COCO format.

\begin{figure*}[t]
\centering
\includegraphics[width=\linewidth]{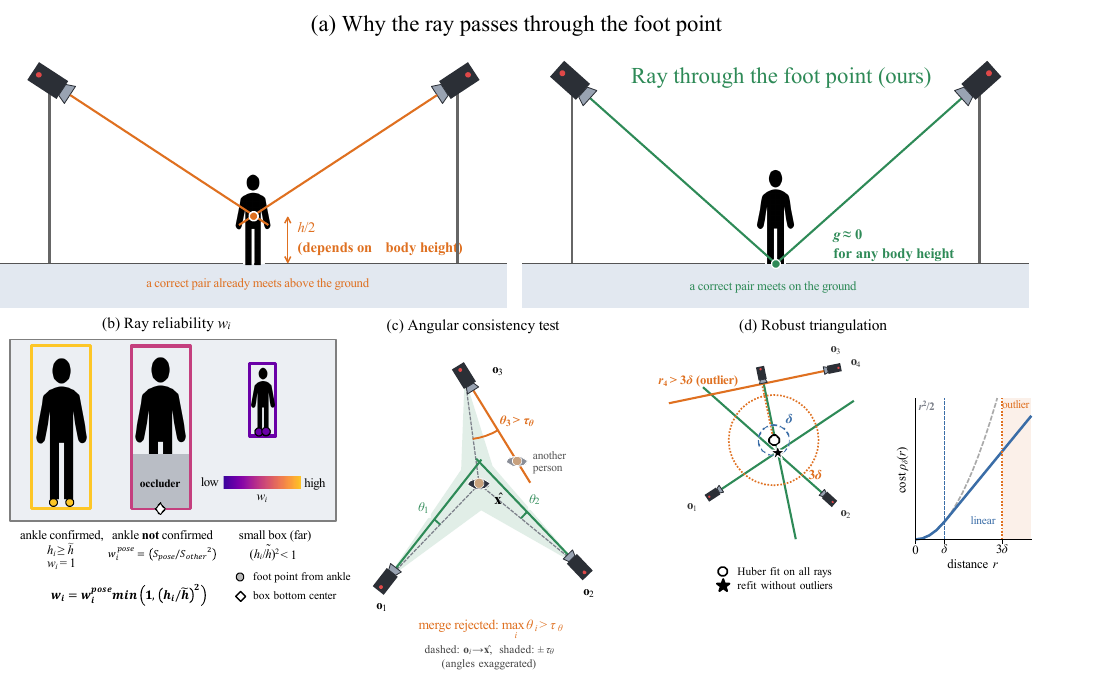}
\caption{Ray construction, reliability, matching, and triangulation.
(a) Rays through the foot point intersect near the ground regardless of body height. Rays through the box center intersect near half the body height even for a correct match.
(b) Ray reliability $w_i$ in Eqs.~\eqref{eq:wpose} and~\eqref{eq:w} decreases when ankle keypoints are not confirmed or the bounding box is small.
(c) A merge is accepted only when every ray satisfies $\theta_i\leq\tau_\theta$ (Eq.~\eqref{eq:consist}). Ray~3 comes from another pedestrian, so $\theta_3>\tau_\theta$ and the merge is rejected. Angles are exaggerated for clarity.
(d) Huber IRLS estimates $\hat{\mathbf{x}}$ (Eq.~\eqref{eq:position}). Rays beyond the outlier threshold are removed before the position is estimated again.}
\label{fig:ray_geometry}
\end{figure*}

When both ankle keypoints satisfy the confidence threshold, the foot point is their mean position; when only one does, we use that ankle.
When neither does, the foot point is the bottom center of the bounding box.
The foot point approximates ground contact, so rays from the same pedestrian intersect near the ground regardless of body height (Fig.~\ref{fig:ray_geometry}a, right).
By contrast, rays through the box center intersect near half the body height even for a correct match (Fig.~\ref{fig:ray_geometry}a, left).
The foot point therefore provides a common reference height across pedestrians.

\begin{figure}[t]
\centering
\includegraphics[width=\linewidth]{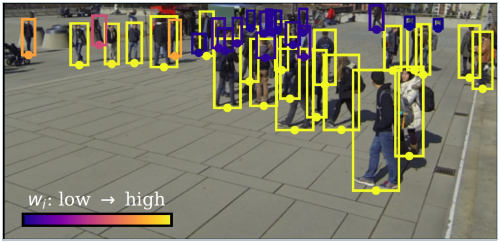}
\caption{Ray reliability $w_i$ on WildTrack frame 1900. Each box is a detection and each dot is the foot point that defines its ray (Sec.~\ref{sec:rays}). Color is $w_i$ from Eq.~\eqref{eq:w}: distant people project to short boxes, so their foot point carries a larger angular error and receives a smaller weight. Triangulation and view support use these weights; no ground truth is involved.}
\label{fig:reliability}
\end{figure}

\paragraph{Angular residual.}

The angular residual between a 3D point $\mathbf{x}$ and the $i$-th ray, with camera center $\mathbf{o}_i$ and unit ray direction $\mathbf{d}_i$, is
\begin{equation}
\label{eq:theta}
\theta_i(\mathbf{x}) =
\arccos\left(\mathbf{d}_i^\top\frac{\mathbf{x}-\mathbf{o}_i}{\|\mathbf{x}-\mathbf{o}_i\|}\right).
\end{equation}
{where $\theta_i(\hat{\mathbf{x}})$ is the residual of ray $i$ at the estimated position $\hat{\mathbf{x}}$.}

\paragraph{Angular error scale.}
\label{sec:sigmatheta}

We estimate $\sigma_\theta$ from each input video in four steps.
First, we form likely same-pedestrian pairs: among the rays of different cameras, we keep only the pairs whose OSNet appearance features are mutual nearest neighbors, and we do not use the association result.
Second, we measure the deviation of each pair: we take the closest point of the two rays (the midpoint of their common perpendicular) and record, for each camera, the angle between the ray direction and the direction to this point, which gives two samples per pair.
Third, we accumulate the samples from the first frame to the current frame $t$.
Fourth, we set $\sigma_\theta(t)$ to $2.097$ times the median of the samples.
The factor $2.097$ is $1.4826\times\sqrt{2}$: $1.4826$ converts the median into a standard deviation for Gaussian errors, and $\sqrt{2}$ accounts for the errors of the two independent rays of a pair. We fixed it before we evaluated accuracy.

\paragraph{Ray reliability.}

Rays differ in accuracy, so each ray $i$ is assigned a reliability weight $w_i\in(0,1]$.
The first cue is ankle visibility: a ray through a confirmed ankle is more accurate than a ray through the box bottom center.
We measure this difference on the input sequence itself: we first estimate $\hat{\mathbf{x}}$ with all weights set to $w_i=1$ and compute the residual $\theta_i(\hat{\mathbf{x}})$ of every ray.
Let $s_{\mathrm{pose}}$ and $s_{\mathrm{other}}$ be the median residuals of the rays with and without a confirmed ankle, taken over all cameras and frames of the input sequence.
The pose-based weight of ray $i$ is
\begin{equation}
\label{eq:wpose}
w_i^{\mathrm{pose}} =
\begin{cases}
1, & \text{if ankle keypoints are confirmed},\\
\left(s_{\mathrm{pose}}/s_{\mathrm{other}}\right)^2, & \text{otherwise}.
\end{cases}
\end{equation}
{The ratio describes the accuracy of rays without a confirmed ankle on the whole sequence, so every such ray receives the same weight.
The square converts the ratio of the residuals, which are standard deviations, into the ratio of the inverse variances, which weights the loss in Eq.~\eqref{eq:position}.}

The second cue is box size: a distant pedestrian has a small bounding box and a less precise foot point.
With the box height $h_i$ and the median box height $\tilde{h}$ of the sequence, the final reliability is
\begin{equation}
\label{eq:w}
w_i =
w_i^{\mathrm{pose}}
\min
\left(
1,
\left(\frac{h_i}{\tilde{h}}\right)^2
\right).
\end{equation}
Fig.~\ref{fig:ray_geometry}(b) shows three typical cases: confirmed ankles, unconfirmed ankles, and a small box for a distant pedestrian.

\subsection{Matching pedestrians across cameras}
\label{sec:matchdetail}

For each frame, detections from different cameras are grouped when they are likely to correspond to the same pedestrian.
Each detection initially forms its own group, and two groups are merged step by step.
Two groups can be merged only if they do not contain detections from the same camera.

For a merge candidate, we first estimate the 3D position $\hat{\mathbf{x}}$ that best fits all rays in the merged group, using the estimator in Sec.~\ref{sec:position_estimation}.
The merge is accepted only when every ray lies within $\tau_\theta$ of the direction from its camera to $\hat{\mathbf{x}}$ (Fig.~\ref{fig:ray_geometry}(c)).
\begin{equation}
\label{eq:consist}
\max_i \theta_i(\hat{\mathbf{x}}) \leq \tau_\theta, {\qquad \tau_\theta = 2\sigma_\theta,}
\end{equation}
where $\tau_\theta$ is {the} angular consistency threshold and $\sigma_\theta$ 
denotes the angular error scale of a single ray.

$\sigma_\theta$ is estimated from each input video (Sec.~\ref{sec:sigmatheta}).
Thus, a merge is rejected if even one ray exceeds $\tau_\theta$.
This prevents one mismatched detection from being hidden by several consistent rays.
Fig.~\ref{fig:ray_geometry}(c) shows this case: one mismatched ray is enough to reject the merge.
{Table~\ref{tab:gate} varies the factor 2 of $\tau_\theta=2\sigma_\theta$.}
In Eq.~\eqref{eq:consist}, $\max_i$ takes the largest angular residual among the rays of the merged group, so the group is judged by its worst ray and not by the average.
Merge candidates are processed in order of the merge cost, the normalized ray distance of the merged group (Sec.~\ref{sec:impldetail}) minus the cosine similarity of the OSNet features, from the smallest cost.
The merge cost only sets the order; the appearance test and the geometric consistency test below decide whether a merge is accepted.
The appearance distance between detections $i$ and $j$ is the cosine distance $1-\mathbf{f}_i^\top\mathbf{f}_j$, where $\mathbf{f}_i$ and $\mathbf{f}_j$ are their L2-normalized OSNet features, and it ranges from 0 to 2.
The distance between two groups is the mean of this distance over all pairs of detections taken from each group.

A merge is rejected when this distance exceeds an appearance threshold $\tau_{\mathrm{app}}$, except in the case described in the next paragraph. The threshold is fixed across all datasets (Table~\ref{tab:constants}), and the experiment that varies it is in Sec.~\ref{sec:sweep}.

When each group contains only one detection, appearance can reject a correct pair.
{We relax the appearance threshold for a pair of two single detections only if the pair fits clearly better than every alternative.}
{For each of the two detections, the pair residual must be at most $\tau_{\mathrm{ratio}}$ times the smallest residual to any other detection in the opposite camera.}
A pair is therefore accepted only when it is clearly better than the best alternative.
If an alternative fits almost as well, the appearance threshold is not relaxed.
The threshold is fixed for all datasets (Table~\ref{tab:constants}), and its sensitivity is reported in Sec.~\ref{sec:sweep}.

Every accepted merge must satisfy the geometric consistency test.

\subsection{Robust 3D position estimation}
\label{sec:robustdetail}

We estimate each pedestrian's 3D position by finding the point that best fits
the rays from multiple cameras.
Each ray is weighted by its reliability $w_i$ from Sec.~\ref{sec:rays}.
Incorrect associations can produce rays that are far from the true position.
We therefore use iteratively reweighted least squares with a Huber loss to
reduce the influence of large errors.

We define the reconstructed scene extent $E$ as the 90th percentile of the Euclidean distances between pairs of points in the reconstructed point cloud after removing low-confidence points.
Distance parameters are defined relative to $E$ because RGB reconstruction has an arbitrary scale.

We fit a 3D position to the rays using their perpendicular distances.
To reduce the effect of a mismatched ray, we use the Huber loss $\rho_\delta(r)$. Errors up to $\delta$ are penalized quadratically, while larger errors are penalized linearly (Fig.~\ref{fig:ray_geometry}(d), right).
The Huber scale $\delta$ is defined relative to $E$ (Table~\ref{tab:constants}).
We estimate the 3D position as
\begin{equation}
\label{eq:position}
\hat{\mathbf{x}}
=
\arg\min_{\mathbf{x}}
\sum_{i\in\mathcal{R}}
w_i
\rho_\delta
\left(
\left\|
(\mathbf{I}-\mathbf{d}_i\mathbf{d}_i^\top)
(\mathbf{x}-\mathbf{o}_i)
\right\|
\right),
\end{equation}
where $\mathcal{R}$ is the set of rays assigned to the pedestrian and the norm is the perpendicular distance $r$ from $\mathbf{x}$ to ray $i$.

We estimate the position freely in 3D rather than forcing it onto the ground.
This is important because incorrect cross-camera matches often produce
intersections above or below the ground, which provides a useful signal for rejecting them in {block \cnum{2} of Fig.~\ref{fig:pipeline} (Sec.~\ref{sec:selection})}.
The final pedestrian location is obtained by projecting the estimated 3D
position onto the ground plane.

\subsection{Candidate selection details}
\label{sec:selectdetail}

{Candidate selection proceeds in four steps. We remove candidates whose rays remain inconsistent, recount the cameras that support each candidate, apply the uncertainty and ground-contact filters (Sec.~\ref{sec:selection}), and remove duplicates by NMS. Candidates are ranked by the score of Sec.~\ref{sec:scoring}.}

After 3D position estimation, we first remove candidates whose rays remain
geometrically inconsistent.
A candidate is removed when the root-mean-square (RMS) angular
error of its rays, $\sqrt{\frac{1}{|\mathcal{R}|}\sum_{i\in\mathcal{R}}\theta_i(\hat{\mathbf{x}})^2}$ with $\theta_i$ from Eq.~\eqref{eq:theta}, exceeds the fixed angular consistency threshold used for association.
The test is repeated after {re-triangulation (Sec.~\ref{sec:retriangdetail})} because it can change the estimated position after association.

Cross-camera association can split detections of the same pedestrian into
multiple candidates.
We therefore recount how many cameras support each estimated 3D position.
For each camera, any detection ray that is close to the position in angle can provide support, even if that detection was originally assigned
to another candidate.
When multiple rays from the same camera support the position, we use the one
with the highest reliability $w_i$.
Allowing this shared support prevents an early association error from removing
the correct pedestrian position.
{The uncertainty filter and the ground-contact filter (block \cnum{2} of Fig.~\ref{fig:pipeline}, Sec.~\ref{sec:selection})}
{then determine whether each candidate is sufficiently reliable.}

\paragraph{Duplicate removal.}

We remove duplicate nearby pedestrian positions using
non-maximum suppression (NMS) with radius $r$ (Sec.~\ref{sec:impldetail}).
When several candidates overlap, those with a higher score are kept first.

\subsection{Trajectory linking details}
\label{sec:trajdetail}

Accepted positions are linked between consecutive frames.
We first keep only pairs within a distance limit.
For these nearby pairs, the matching cost combines ground-plane distance and OSNet appearance distance, and the Hungarian algorithm selects one-to-one links with the lowest costs.
The distance limit prevents large jumps, while appearance distinguishes pedestrians that are close to each other.

\section{Candidate Refinement}
\label{sec:refinementdetail}

{This section describes the candidate refinement of Sec.~\ref{sec:refinement}, which consists of re-triangulation in block~\cnum{1} and the score-based recheck in block~\cnum{2}. Their hyperparameters are listed in Sec.~\ref{sec:hyperparams}.}

\subsection{Re-triangulation}
\label{sec:retriangdetail}

{Re-triangulation (Sec.~\ref{sec:refinement}) starts from the robust estimate of Sec.~\ref{sec:robustdetail}.}
After the initial estimate,
rays whose distance from the estimated position exceeds $\kappa_\delta\delta$ are treated as outliers and removed (Fig.~\ref{fig:ray_geometry}(d), left).
If at least one outlier is found and at least two rays are within the cutoff, we estimate the
position again using only the rays within the cutoff.
This prevents rejected rays from shifting the final position.
Fig.~\ref{fig:ray_geometry}(d) shows both the Huber loss and the re-estimation after the removal of an outlier.
{Table~\ref{tab:retri} lists the settings of this re-triangulation.}

\subsection{Score-based recheck}
\label{sec:scoring}

{The score implements the recheck of Sec.~\ref{sec:refinement}.}
Each candidate gets seven image-derived cues: number of supporting rays, negative median residual, detector confidence, view support ratio, consistency of implied person heights, temporal persistence, and reconstruction confidence at the ray pixels {(Table~\ref{tab:signals}). All cues are oriented so that a larger value is better.}
The cues are standardized over the sequence and weighted by their Otsu separability.
{Standardization is the $z$-score. The Otsu separability $\omega_k$ of a cue is the ratio of the between-class variance of its Otsu split to its total variance. The six cues other than reconstruction confidence are combined into}
\begin{equation}
\label{eq:score6}
S_6=\sum_{k=1}^{6}\omega_k z_k ,
\end{equation}
{where $z_k$ is the standardized value of cue $k$.}
Reconstruction confidence $d$ is also scaled by its correlation with the consensus of the other six cues{, which is $S_6$ of Eq.~\eqref{eq:score6}. Its gain is $\alpha=|\mathrm{corr}(d,S_6)|$ over all candidates, and the final score is}
\begin{equation}
\label{eq:score}
S=S_6+\alpha\,\omega_d\,z_d ,
\end{equation}
{where $\omega_d$ and $z_d$ are the Otsu separability and the standardized value of $d$.}
The combined score $S$ of Eq.~\eqref{eq:score} ranks candidates for NMS. {Candidates at or above the Otsu threshold are accepted, and the remaining candidates are added in order of score unless they lie within the NMS radius $r$ of an accepted candidate.}

{The view support ratio uses an angular tolerance $\tau_{\mathrm{view}}$ that is estimated from each sequence. For candidate $i$ in frame $f_i$ and camera $c$, let $\psi_{ic}=\min_{r}\arccos\langle\hat{\mathbf{h}}_r,\hat{\mathbf{u}}_{ic}\rangle$ over the rays $r$ of camera $c$ in frame $f_i$, where $\hat{\mathbf{h}}_r$ is the ground direction of ray $r$ and $\hat{\mathbf{u}}_{ic}$ is the ground direction from camera $c$ to candidate $i$. Let $\mathcal{C}(i)$ be the cameras that formed candidate $i$, and $\Theta=\{\log_{10}\psi_{ic}: c\notin\mathcal{C}(i)\}$. We split the histogram of $\Theta$ by the Otsu criterion,}
\begin{equation}
\label{eq:tauview}
\tau_{\mathrm{view}}=10^{s^{\ast}},\qquad \\s^{\ast}=\arg\max_{s}\,p_0(s)\,p_1(s)\,\bigl(m_0(s)-m_1(s)\bigr)^2,
\end{equation}
{where $p_0$, $p_1$ and $m_0$, $m_1$ are the weights and the means of the two classes separated at $s$.}

\begin{table*}[t]
\centering
\caption{Cues of the candidate score for a candidate with rays $\mathcal{R}$ and estimated position $\hat{\mathbf{x}}$ (Eq.~\eqref{eq:position}). The residual and the height variation are better when small, so they carry a minus sign; a larger value is better for every cue.}
\label{tab:signals}
\footnotesize
\setlength{\tabcolsep}{4pt}
\begin{tabular}{@{}p{2.9cm}p{4.6cm}p{8.8cm}@{}}
\toprule
Cue & Definition & Notation \\
\midrule
Supporting rays & $|\mathcal{R}_{\mathrm{in}}|$ & $\mathcal{R}_{\mathrm{in}}$: rays judged to be inliers (Sec.~\ref{sec:robustdetail}) \\
Median residual & $-\operatorname{median}_{i\in\mathcal{R}}\, e_i(\hat{\mathbf{x}})$ & $e_i(\mathbf{x})=\|(\mathbf{I}-\mathbf{d}_i\mathbf{d}_i^\top)(\mathbf{x}-\mathbf{o}_i)\|$, the perpendicular distance in Eq.~\eqref{eq:position}; $\hat{\mathbf{x}}$ is re-estimated from the inliers \\
Detector confidence & $\dfrac{1}{|\mathcal{R}|}\sum_{i\in\mathcal{R}} s_i$ & $s_i$: detection score of ray $i$ (Sec.~\ref{sec:candidates}) \\
View support ratio & $\dfrac{1}{C}\sum_{c=1}^{C} m_c$ & $m_c=1$ if $\psi_{ic}<\tau_{\mathrm{view}}$ (Eq.~\eqref{eq:tauview}), and $0$ otherwise; $\psi_{ic}$: smallest azimuth difference between the rays of camera $c$ and the direction to candidate $i$ \\
Height consistency & $-\sigma_h/\mu_h$ & $h_i=\dfrac{L_i}{f_i}\,p_i\,\gamma$ with $L_i=\|\hat{\mathbf{x}}-\mathbf{o}_i\|$, focal length $f_i$, box height $p_i$ in pixels (feet to head), and image scale $\gamma$; $\mu_h$ and $\sigma_h$ are the mean and standard deviation of $h_i$ over $\mathcal{R}$ \\
Temporal persistence & $\dfrac{1}{2}\,(n_{t-1}+n_{t+1})$ & $n_{t'}=1$ if frame $t'$ has another candidate within the NMS radius $r$ of $\hat{\mathbf{x}}$ (Sec.~\ref{sec:impldetail}), and $0$ otherwise \\
Reconstruction confidence & $\dfrac{1}{|\mathcal{R}|}\sum_{i\in\mathcal{R}} q(\mathbf{u}_i)$ & $q$: VGGT-$\Omega$ depth confidence map of the initial frame (Sec.~\ref{sec:cameradetail}); $\mathbf{u}_i$: pixel of ray $i$ \\
\bottomrule
\end{tabular}
\end{table*}

\section{Hyperparameters}
\label{sec:hyperparams}

{Table~\ref{tab:constants} lists the main fixed settings of the pipeline of Sec.~\ref{sec:method}, and Table~\ref{tab:retri} lists those of re-triangulation in the candidate refinement of Sec.~\ref{sec:refinement}.}

\subsection{Implementation settings and scale normalization}
\label{sec:impldetail}

\paragraph{Implementation settings.}

All fixed parameters are shared across datasets, and Table~\ref{tab:constants} lists the main ones. Camera refinement uses pedestrians supported by at least three cameras.The same procedure is used for every dataset: we sample up to 4{,}000 valid reconstructed points (4{,}000 is an upper limit, and all points are used if a scene has fewer), then sample up to 200{,}000 of the pairs of these points with a fixed random seed, and take the 90th percentile of their distances as $E$.

\paragraph{Scale normalization.}

Distance thresholds depend on the unknown reconstruction scale, so we normalize every reconstructed distance $d$ by the scene extent $\ext$.
\begin{equation}
\tilde d = d/\ext.
\end{equation}
{where $\ext$ is the 90th percentile of the distances $\|\mathbf{p}_a-\mathbf{p}_b\|$ between pairs of reconstructed points $\mathbf{p}_a$ and $\mathbf{p}_b$ (Sec.~\ref{sec:robustdetail}).}
Equivalently, a threshold written as $c\ext$ uses the same normalized threshold $c$ across scenes; for example, the Huber scale is $\delta=0.01\ext$ (Table~\ref{tab:constants}) and the NMS radius is $r=0.018\ext$. Angular errors are scale-independent, so $\sigt$ needs no normalization by $\ext$ and is estimated from each video (Sec.~\ref{sec:sigmatheta}).

\begin{table}[t]
\centering
\caption{Settings of re-triangulation. The outlier cutoff is defined relative to the Huber scale $\delta$ (Table~\ref{tab:constants}).}
\label{tab:retri}
\footnotesize
\begin{tabular}{@{}lp{3.9cm}@{}}
\toprule
Setting & Value \\
\midrule
Outlier cutoff $\kappa_\delta\delta$ & $3\delta$ ($\kappa_\delta=3$) \\
Re-estimation condition & at least one outlier and at least two rays within the cutoff \\
\bottomrule
\end{tabular}
\end{table}

\begin{table}[t]
\centering
\caption{Fixed settings shared across all datasets. $\ext$ is the reconstructed scene extent. {Table~\ref{tab:svc} varies the similarity threshold of single-view completion.}}
\label{tab:constants}
\footnotesize
\begin{tabular}{@{}p{2.3cm}p{5.6cm}@{}}
\toprule
Step & Setting \\
\midrule
Rays & ankle confidence $\ge 0.25$ \\
Association & appearance threshold 0.6; uniqueness ratio 0.3 \\
Triangulation & Huber scale $\delta=0.01\ext$ \\
Single-view completion & OSNet cosine similarity $> 0.9$ for joining single-camera track fragments \\
\bottomrule
\end{tabular}
\end{table}

\section{Experimental Details}
\label{sec:expdetails}

This section collects the evaluation protocol and the datasets that Sec.~\ref{sec:results} refers to.

\subsection{Evaluation protocol}
\label{sec:evaldetail}

The reconstructed coordinate system differs from each dataset's coordinate system by a global rotation, translation, and scale.
For evaluation only, we align the predictions with a 7-DoF Umeyama transform~\cite{umeyama1991} fitted between the estimated and ground-truth camera centers.
Because the alignment uses ground-truth camera positions, the reported results measure relative localization rather than absolute metric localization.
The metrics use the multi-view detection evaluator, which ignores predictions outside the annotated evaluation area.

\subsection{Datasets and sequences}
\label{sec:datasetdetail}

\textbf{WildTrack}~\cite{chavdarova2018wildtrack} contains 7 real cameras.
We evaluate the standard test frames 1800--1995, sampling every fifth frame for 40 frames and 952 pedestrian instances.
We also evaluate the 360 annotated frames 0--1795, containing approximately 8{,}570 instances.

\noindent \textbf{MultiviewX}~\cite{hou2020mvdet} contains 6 synthetic cameras.
We evaluate frames 360--399, containing 40 frames and approximately 1{,}490 pedestrian instances.

\noindent \textbf{GMVD}~\cite{vora2023gmvd} contains multiple synthetic scenes.
GMVD has seven scenes; its protocol trains supervised methods on six scenes and tests them on the remaining one, the test scene. CAT-Free needs no training, so we use only the test scene, and our results can be compared with methods trained on the other six scenes. The test scene has ten videos: five recorded with a 6-camera layout and five with an 8-camera layout.
Our GMVD result is the evaluation across all ten videos of the test scene.
Below, we call the two camera layouts layout 1 (6 cameras) and layout 2 (8 cameras), and number the videos of each layout seq1, seq2, and so on.
Camera geometry of each video is estimated from its RGB images, while pedestrian annotations are used only after inference to compute metrics.

\section{Additional Results}
\label{sec:gmvdmore}

On GMVD, CAT-Free achieves 76.1 MODA and 82.1 IDF1 (Table~\ref{tab:ablation}, row (e)).
The filter, threshold, and candidate analyses below use the multi-view pipeline.

\begin{table*}[t]
\centering
\caption{All metrics of the ablation in Table~\ref{tab:ablation}. Each row adds one component to the row above. Rows ``+ reliability'', ``+ uncertainty filter'', ``+ ground-contact filter'', ``+ candidate refinement'', and ``+ SVC'' correspond to rows (a)--(e) of Table~\ref{tab:ablation}; ``core'' is the same candidates without the ray reliability weights. MODA and MODP use a 0.5\,m matching threshold, and MOTA and IDF1 use 1\,m. The best value in each column is in bold.}
\label{tab:ablation_full}
\footnotesize
\setlength{\tabcolsep}{3.5pt}
\begin{tabular}{@{}l rrr rrr rrr rrr@{}}
\toprule
& \multicolumn{3}{c}{MODA} & \multicolumn{3}{c}{MODP} & \multicolumn{3}{c}{MOTA} & \multicolumn{3}{c}{IDF1} \\
\cmidrule(lr){2-4}\cmidrule(lr){5-7}\cmidrule(lr){8-10}\cmidrule(lr){11-13}
& WT & MVX & GMVD & WT & MVX & GMVD & WT & MVX & GMVD & WT & MVX & GMVD \\
\midrule
core & 69.1 & 71.4 & 71.5 & 73.8 & 74.7 & 63.1 & 65.1 & 63.7 & 69.1 & 56.3 & 52.9 & 58.4 \\
+ reliability & 78.9 & 73.0 & 67.9 & \textbf{75.1} & \textbf{75.7} & \textbf{64.7} & 74.2 & 65.7 & 65.4 & 56.3 & 55.5 & 59.6 \\
+ uncertainty filter & 76.5 & 73.4 & 71.9 & 74.3 & 75.4 & 63.7 & 71.7 & 65.7 & 69.4 & 58.9 & 56.0 & 60.3 \\
+ ground-contact filter & 80.3 & 75.0 & 65.3 & 74.4 & 75.6 & 64.1 & 75.5 & 67.3 & 61.7 & 60.0 & 57.8 & 53.9 \\
+ candidate refinement & 87.5 & 82.8 & 75.2 & 73.1 & 73.4 & 63.0 & 85.6 & 77.6 & 73.6 & 71.9 & 64.2 & 61.0 \\
+ SVC & \textbf{89.2} & \textbf{84.1} & \textbf{76.1} & 73.0 & 73.1 & 62.6 & \textbf{90.8} & \textbf{82.7} & \textbf{78.2} & \textbf{90.4} & \textbf{81.6} & \textbf{82.1} \\
\bottomrule
\end{tabular}
\end{table*}

\subsection{Camera parameters: direct output or back-computation}
\label{sec:vggtcam}

\begin{figure}[t]
\centering
\includegraphics[width=\columnwidth]{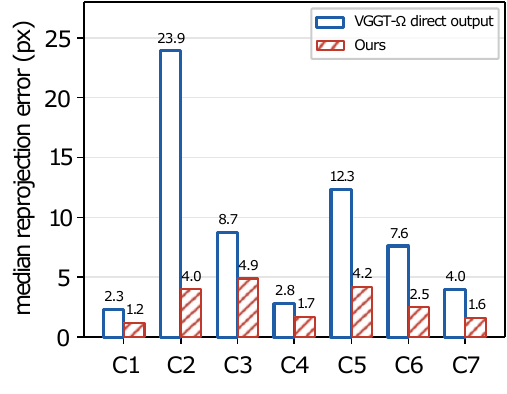}
\caption{Median reprojection error of the reconstructed 3D points in each WildTrack camera (C1--C7), for the camera parameters that VGGT-$\Omega$ outputs directly and for those back-computed from the point cloud by {Perspective-n-Point (PnP) estimation} 
(lower means more accurate).}
\label{fig:camera_source}
\end{figure}

Sec.~\ref{sec:candidates} back-computes the camera parameters from the point cloud of VGGT-$\Omega$ and does not use the camera parameters that VGGT-$\Omega$ outputs directly.
For each of the seven WildTrack cameras, we project the reconstructed 3D points into the image with each set of camera parameters and measure the median distance (reprojection error) to the pixels that the points belong to.
The points are reconstructed from the images of that camera, so camera parameters closer to the true ones give a smaller error.
Fig.~\ref{fig:camera_source} shows that the back-computed camera parameters have a smaller reprojection error than the directly output ones in all seven cameras, and the gap is largest in camera~2, where the error falls from about 24\,px to about 4\,px.
This indicates that the back-computed camera parameters are more accurate.

\subsection{3D reconstruction backbone}
\label{sec:backbone}

We replace VGGT-$\Omega$ by other feed-forward 3D reconstruction models, namely Pi3, Pi3X~\cite{wang2026pi3}, DA3~\cite{lin2026da3} (Large and Nested-Giant-Large), and MUSt3R~\cite{cabon2025must3r}, and keep everything else unchanged, to assess how much the choice of the reconstruction model matters.
Table~\ref{tab:backbone} shows that VGGT-$\Omega$ outperforms the other models on WildTrack in MODA, MOTA, and IDF1, by more than 10 percentage points in MODA.
Pi3X outperforms VGGT-$\Omega$ only in MODP and MOTP.
This result supports the choice of VGGT-$\Omega$ as the reconstruction model.

\begin{table}[t]
\centering
\caption{WildTrack results with different 3D reconstruction models. MOTP is a distance, so lower is better. Bold marks the best value in each column.}
\label{tab:backbone}
\footnotesize
\setlength{\tabcolsep}{4pt}
\begin{tabular}{@{}l rrrrr@{}}
\toprule
Model & MODA & MODP & MOTA & MOTP$\downarrow$ & IDF1 \\
\midrule
VGGT-$\Omega$ & \textbf{89.18} & 73.01 & \textbf{90.76} & 0.157 & \textbf{90.35} \\
Pi3 & 75.63 & 69.23 & 77.21 & 0.17 & 83.81 \\
Pi3X & 73.74 & \textbf{74.50} & 73.21 & \textbf{0.14} & 82.06 \\
DA3-Large & 76.68 & 69.03 & 79.20 & 0.17 & 81.67 \\
DA3-Nested-Giant-Large & 78.99 & 63.81 & 78.78 & 0.19 & 83.21 \\
MUSt3R & 45.90 & 29.67 & 75.84 & 0.41 & 81.45 \\
\bottomrule
\end{tabular}
\end{table}

\subsection{YOLO pose model}
\label{sec:detectorpose}


{We replace the YOLO pose model, which gives both the person detections and the keypoints, to check how sensitive CAT-Free is to this model.
Table~\ref{tab:detector} shows that the control, YOLO11x-pose, outperforms the four other models on WildTrack on every metric.
Among the four, model size and generation have little effect.
This indicates that CAT-Free retains most of its accuracy with other YOLO pose models.}

\begin{table}[t]
\centering
\caption{WildTrack results with different YOLO pose models, each of which gives both the person detections and the keypoints. The control is YOLO11x-pose. MOTP is a distance, so lower is better. Bold marks the best value in each column.}
\label{tab:detector}
\footnotesize
\setlength{\tabcolsep}{4pt}
\begin{tabular}{@{}l rrrrr@{}}
\toprule
Model & MODA & MODP & MOTA & MOTP$\downarrow$ & IDF1 \\
\midrule
YOLO11x-pose & \textbf{89.18} & \textbf{73.01} & \textbf{90.76} & \textbf{0.157} & \textbf{90.35} \\
YOLO26x-pose & 80.78 & 69.50 & 83.19 & 0.164 & 84.95 \\
YOLO26n-pose & 81.83 & 68.99 & 82.98 & 0.171 & 85.36 \\
YOLO11s-pose & 80.99 & 69.59 & 82.25 & 0.170 & 84.40 \\
YOLO11n-pose & 80.78 & 69.42 & 82.04 & 0.179 & 85.16 \\
\bottomrule
\end{tabular}
\end{table}

\subsection{Per-sequence versus fixed thresholds}

{Table~\ref{tab:quantile} replaces the two log-Otsu thresholds by one fixed quantile of the candidates, the same across datasets.}

\begin{table*}[t]
\centering
\caption{Per-sequence versus fixed thresholds (MODA). Both filter thresholds are replaced by the same fixed quantile on every dataset. {Log-Otsu has the highest mean. The best fixed quantile differs by dataset.}}
\label{tab:quantile}
\footnotesize
\begin{tabular}{@{}lrrrr@{}}
\toprule
Threshold rule & WildTrack & MultiviewX & GMVD & Mean \\
\midrule
fixed quantile p50 & {84.2} & {59.9} & {46.4} & {63.5} \\
fixed quantile p60 & {86.7} & {74.4} & {61.9} & {74.3} \\
fixed quantile p70 & {84.6} & {\textbf{84.2}} & {75.4} & {81.4} \\
log-Otsu (default) & {\textbf{89.2}} & {84.1} & {\textbf{76.1}} & {\textbf{83.1}} \\
\bottomrule
\end{tabular}
\end{table*}

{Table~\ref{tab:quantile} shows that log-Otsu outperforms the best fixed quantile, p70, in mean MODA.
On WildTrack and GMVD, log-Otsu outperforms every fixed quantile, and on MultiviewX it is comparable to p70.
The best fixed quantile differs by dataset, p60 on WildTrack and p70 on MultiviewX and GMVD.
This suggests that no single fixed quantile fits all datasets, and the per-sequence threshold avoids this choice because it adapts to the distribution of each video.}

\subsection{Parameter sensitivity}
\label{sec:sweep}

{We vary the appearance threshold from 0.4 to 0.8, the uniqueness ratio from 0.1 to 1.0, and the outlier cutoff from $1\delta$ to $5\delta$, one constant at a time, on the full splits of WildTrack, MultiviewX, and GMVD.
The frozen values are at or near the best.
Smaller values of the appearance threshold and the outlier cutoff degrade the mean MODA, whereas the uniqueness ratio changes it only slightly.
This indicates that the exact values of these constants have little effect on the accuracy within the tested ranges.
The comparisons above apply to the multi-view pipeline.}

{We also vary the multiplier $\gamma$ of the angular consistency threshold $\tau_\theta=\gamma\sigt$ (Eq.~\eqref{eq:consist}) from 1.5 to 3 on the same splits.
Table~\ref{tab:gate} shows that the frozen value $\gamma=2$ outperforms the larger multipliers in mean MODA and is comparable to $\gamma=1.5$.
A looser threshold degrades MODA on all three datasets, most on WildTrack.
This indicates that the threshold performs well for $\gamma$ from $1.5$ to $2$.}

\begin{table}[t]
\centering
\caption{Ablation of the multiplier $\gamma$ of the angular consistency threshold $\tau_\theta=\gamma\sigt$. The table reports MODA (\%) of CAT-Free. The row $\gamma=2$ is the frozen value and equals row (e) of Table~\ref{tab:ablation}. GMVD is the mean over its ten videos, and Mean is the mean of the three datasets. The best value in each column is in bold.}
\label{tab:gate}
\footnotesize
\begin{tabular}{@{}lrrrr@{}}
\toprule
$\gamma$ & WildTrack & MultiviewX & GMVD & Mean \\
\midrule
1.5 & \textbf{89.39} & 83.40 & \textbf{76.42} & 83.07 \\
2.0 & 89.18 & \textbf{84.07} & 76.10 & \textbf{83.12} \\
2.5 & 87.92 & 83.40 & 75.78 & 82.37 \\
3.0 & 84.14 & 82.53 & 75.26 & 80.64 \\
\bottomrule
\end{tabular}
\end{table}

{We also vary the OSNet cosine similarity threshold for joining single-camera track fragments in SVC from 0.7 to 0.95.
Table~\ref{tab:svc} shows that the frozen value 0.9 outperforms the other thresholds in both mean MODA and mean IDF1.
Every threshold gains more than 15 points in the mean IDF1 over the value without SVC (row (d) of Table~\ref{tab:ablation}).
This indicates that the gain of SVC holds for every tested threshold.}

\begin{table}[t]
\centering
\caption{Ablation of the OSNet cosine similarity threshold for joining single-camera track fragments in SVC. The table reports the mean over the three datasets. The row 0.9 is the frozen value and equals row (e) of Table~\ref{tab:ablation}. The best value in each column is in bold.}
\label{tab:svc}
\footnotesize
\begin{tabular}{@{}lrr@{}}
\toprule
Threshold & MODA & IDF1 \\
\midrule
0.7 & 82.98 & 83.84 \\
0.8 & 82.42 & 81.00 \\
0.9 & \textbf{83.12} & \textbf{84.68} \\
0.95 & 82.60 & 82.56 \\
\bottomrule
\end{tabular}
\end{table}

\subsection{Ground-contact filter in the GMVD videos}
\label{sec:analysis}

{We measure coverage as the fraction of annotated people with at least one candidate within 0.5\,m; ground truth is used only to label outcomes.
Most of the coverage loss of the ground-contact filter on GMVD comes from two of the ten videos (seq2 and seq3 of layout 1), where coverage declines sharply after the filter, and the same two videos explain the MODA drop of the filter on GMVD (Table~\ref{tab:ablation}).
Table~\ref{tab:gmvdvideos} shows that in these two videos, the log-Otsu threshold $\tau_g$ of the filter is much smaller than in the other eight.
At these thresholds, more than half of the feet of the annotated pedestrians lie outside, whereas the best threshold of each video, found with the ground truth, is about twice as large.
With the ground-truth ground plane, the threshold rises, the coverage recovers, and MODA improves.
The five videos of layout 1 share the camera error, and only these two fail.
This indicates that the loss comes from the threshold of the filter, and a more robust threshold would remove it.}

\begin{table}[t]
\centering
\caption{Ground-contact filter in the ten GMVD videos. The coverage drop is the loss of coverage caused by the ground-contact filter, in points, and $\tau_g$ is the log-Otsu threshold of the filter.}
\label{tab:gmvdvideos}
\footnotesize
\begin{tabular}{@{}l rr@{}}
\toprule
Video & Coverage drop & $\tau_g$ (cm) \\
\midrule
Layout 1, seq1 & 2.1 & 22.4 \\
Layout 1, seq2 & 31.0 & 8.8 \\
Layout 1, seq3 & 43.6 & 7.4 \\
Layout 1, seq4 & 1.2 & 25.2 \\
Layout 1, seq5 & 3.4 & 21.8 \\
Layout 2, seq1 & 0.7 & 22.5 \\
Layout 2, seq2 & 0.9 & 25.7 \\
Layout 2, seq3 & 1.1 & 22.3 \\
Layout 2, seq4 & 0.8 & 24.8 \\
Layout 2, seq5 & 0.7 & 23.1 \\
\bottomrule
\end{tabular}
\end{table}

\subsection{Computational Cost}

All steps run on one workstation (NVIDIA RTX A6000) and reuse cached detections, rig,
and reconstruction, so the times below exclude the VGGT-$\Omega$ reconstruction and the camera estimation. Processing from detection through tracking takes 135\,s on WildTrack (40 frames, 7 cameras),
237\,s on MultiviewX (40 frames, 6 cameras), and 158--230\,s {per GMVD video (about 100 frames, 6 or 8 cameras)}.
Cross-view association dominates at 61--71\% of that time. It compares every pair of detections in a frame
and is the only step that scales quadratically with the number of detections per frame.



\end{document}